%% file: main.tex
\documentclass{egpubl}
\usepackage{pg2026}

\SpecialIssueSubmission    

\usepackage[T1]{fontenc}
\usepackage{dfadobe}  
\usepackage{amsmath}
\usepackage{amssymb}
\usepackage{booktabs}
\usepackage{tabularx}
\usepackage{xspace}
\usepackage{mathtools}
\usepackage{mathtools}
\usepackage{xcolor}
\usepackage{enumitem}

\usepackage{cite}  
\BibtexOrBiblatex
\electronicVersion
\PrintedOrElectronic
\ifpdf \usepackage[pdftex]{graphicx} \pdfcompresslevel=9
\else \usepackage[dvips]{graphicx} \fi

\usepackage{egweblnk} 

\input{preamble}
\title{\methodname: Gaussian Splat Completion with 2D Diffusion Priors} 

\author[E. Brugger \& P. Erler \& S. Ohrhallinger \& P. Guerrero]
{\parbox{\textwidth}{\centering 
E. Brugger$^{1}$\orcid{0009-0003-0866-9489} 
and P. Erler$^{1}$\orcid{0000-0002-2790-9279}
and S. Ohrhallinger$^{1}$\orcid{0000-0002-2526-7700}
and P. Guerrero$^{2}$\orcid{0000-0002-7568-2849}
}
        \\
{\parbox{\textwidth}{\centering 
$^1$TU Wien, Austria\\
$^2$Adobe Research, United Kingdom
       }
}
}

\begin{document}

\teaser{
 \includegraphics[width=1.0\linewidth]{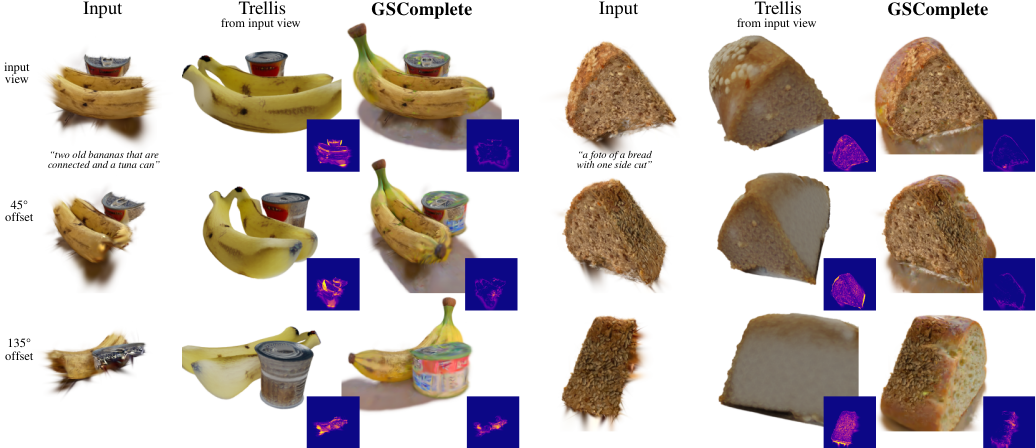}
 \centering
  \caption{We propose \methodname to generatively complete a partial 3D object represented as a set of Gaussian splats. Our approach generates a plausible completed 3D object that exactly preserves the given partial input using only a 2D diffusion prior. Insets show errors in the preservation of input splats for each view.
}
\label{fig:teaser}
}

\maketitle
\thispagestyle{empty}
\pagestyle{empty}

\input{sec/0_abstract}

\input{sec/1_intro}
\input{sec/2_related_work}

\input{sec/3_method}
\input{sec/4_results}
\input{sec/5_conclusion}
\input{sec/6_acknowledgements}



\bibliographystyle{eg-alpha-doi} 
\bibliography{main}       

\end{document}

%% file: preamble.tex
\newcommand{\paul}[1]{\textcolor{orange}{[PAUL: #1]}}

\newcommand{\philipp}[1]{\textcolor{cyan}{[PHILIPP: #1]}}

\newcommand{\methodname}{\text{GSComplete}\xspace}

\newcommand{\PartialSplats}{P}
\newcommand{\FullSplats}{O}
\newcommand{\OneSplatPartial}{p}
\newcommand{\OneSplatFull}{o}

%% file: sec/0_abstract.tex
\begin{abstract}
Gaussian splats provide a fast, high-fidelity representation for 3D objects but are often constructed from incomplete input data in practice, leaving missing regions. Existing completion methods either do not preserve the original splats or require scarcely available 3D training data. We propose GSComplete, which combines 3D generation based on Score Distillation Sampling with a novel preservation loss that encourages the original splats to be preserved where they should be visible. This effectively completes the Gaussian splat object using only 2D diffusion priors while fully preserving existing splats and generating new splats only in missing regions, without occluding the input. To evaluate our approach, we introduce a new dataset of partial Gaussian splat objects
and show that GSComplete achieves significantly more accurate preservation of the input than existing methods with comparable plausibility of the completed result.
Our code and dataset will be made available upon acceptance.

\ccsdesc[500]{Computing methodologies~Shape modeling}

\printccsdesc

\end{abstract}

%% file: sec/1_intro.tex
\section{Introduction}

Gaussian splats provide a fast and high-fidelity representation for 3D objects that can be constructed from image sets, 3D scans, or existing 3D objects. Gaussian splat scenes used in practice are often incomplete due to missing information at construction time, such as incomplete image sets, scan shadows, or partial 3D objects.
For example, a robotic agent or scanner may only be able to view the front of an object, or a 3D artist may want to save time by only modeling the important parts of an object. To complete these objects, generative methods have been proposed that extrapolate missing information, such as novel view synthesis methods~\cite{wu2024reconfusion, liu2023zero, shi2023zero123plus} and native 3D generation methods that employ a 3D prior~\cite{du2025superpc, xiang2025trellis}. However, a 3D prior requires scarcely available 3D training data, and novel view synthesis methods represent existing splats with one or multiple images,
making it difficult to accurately preserve them. Analogously to image completion, we want a Gaussian splat completion method that fully preserves existing splats and only generates new splats in missing regions as needed.

We propose \methodname to complete a Gaussian splat object using only 2D diffusion priors while fully preserving existing splats. Given a set of Gaussian splats that represent a partial 3D object (obtained from sources such as partial scans, image reconstructions, or partial 3D meshes),
our method learns to output a set of Gaussian splats that (i) exactly preserves the original splats, (ii) preserves the appearance of the original object from a given range of viewpoints, making sure it is not hidden by new splats; and (iii) represents a plausible completion of the object.

The central challenge in our problem is that the guidance from the 2D diffusion prior and preservation of the existing object part are at odds to some extent. Typically, slight variations of the existing object part will have a higher probability in the prior than its exact preservation, so the prior will often encourage new splats to cover parts of the existing object. Our solution is to combine a generative approach based on Score Distillation Sampling (SDS)~\cite{poole2022dreamfusion, shi2023mvdream} with a new preservation loss that encourages the original object to be preserved.





To evaluate our approach, we introduce a new dataset of partial Gaussian splat (GS) objects and show that \methodname can achieve significantly more accurate preservation of the original object part than existing methods with comparable plausibility of the completed result.
Our main contributions are: (i) a new approach for GS completion that uses 2D diffusion priors and fully preserves existing splats; and (ii) a new dataset of partial GS splat objects. 










%% file: sec/2_related_work.tex
\section{Related work}

The completion of missing data in point clouds, often due to artifacts from scans such as scan shadows, occlusions, or incomplete campaign planning, is a topic that has been researched for a long time, with input data as point clouds, or single or few images, more recently, using 3DGS - see here for a detailed survey on point cloud completion \cite{zhuang2024survey}.

\paragraph*{Point cloud completion with 3D priors.} This approach trains a model specifically for 3D point cloud completion or 3D point cloud generation. It requires scarcely available 3D data, and as a result, these methods typically do not generalize beyond their limited training data. One of the first learning-based methods is Point Completion Network~\cite{yuan2018pcn}, AdaPoinTr~\cite{yu2301adapointr} uses transformers with a denoising task, 3D-PCGR~\cite{yuan20243d} combines LIDAR and color data, colorizing using a GAN, and recent SuperPC~\cite{du2025superpc} proposes a unified diffusion framework for completing, upsampling, denoising, and colorizing point clouds.

\paragraph*{Single- or few-image 3D reconstruction.} An alternative approach to complete 3D Gaussian splats is to render existing splats from one or multiple viewpoints that do not show the missing parts, and reconstruct the full object from these images. However, as the existing splats are represented by one or a few images, this does not accurately preserve existing splats, as we demonstrate in Section~\ref{sec:results}. Tewari et al.~\cite{tewari2023diffusion} use a differentiable forward renderer for the denoising of a conditional diffusion model, avoiding 3D supervision. Splatter Image~\cite{szymanowicz2024splatter_image} turns each pixel into a splat. Shen et al.~\cite{shen2024pixel} extend it to hierarchically map pixels to multiple splats. DiffGS~\cite{zhou2024diffgs} proposes a continuous Gaussian Splatting function for unstructured 3DGS.  pixelSplat~\cite{charatan2024pixelsplat} samples Gaussian means from a dense probability distribution. LGM~\cite{tang2024lgm} produces multi-view Gaussian features with an asymmetric U-Net. MVSplat~\cite{chen2024mvsplat} stores cross-view feature similarities in a cost volume. LRM~\cite{hong2024lrm}, GS-LRM~\cite{gslrm2024}, and TripoSR~\cite{TripoSR2024} incorporate transformers. Wonder3D~\cite{long2024wonder3d} applies cross-domain diffusion. Direct3D~\cite{wu2024direct3d} uses a native 3D generative model, and Wu et al.~\cite{wu2025direct} extend it with spatial sparse attention. ReconFusion~\cite{wu2024reconfusion} uses NeRFs. latentSplat~\cite{wewer2024latentsplat} predicts semantic Gaussians in a 3D latent space and decodes them quickly in 2D. Flash3D~\cite{szymanowicz2025flash3d} creates a layer of Gaussians at the predicted depth and further layers behind that. Ouroboros3D~\cite{wen2025ouroboros3d} uses a recursive diffusion-reconstruction-based process. Trellis~\cite{xiang2025trellis} generates native 3D built on a unified structured latent representation and rectified flow transformers.

\paragraph*{3D inpainting with 2D priors.} This approach directly inpaints missing splats, using a 2D inpainting prior that is projected back to the 3D scene. The difference to completion is that this usually removes objects from a background and only completes missing parts that are small compared to the size of the full scene or object, such as holes or small scan shadows: Weder et al.~\cite{Weder2023Removing}, Zhang et al.~\cite{zhang2023point}, RGBD2~\cite{lei2023rgbd2}, OR-NeRF~\cite{yin2023ornerf}, Gaussian Grouping~\cite{ye2024gaussian}, MVIP-NeRF~\cite{chen2024mvip}, MALD-NeRF~\cite{lin2024maldnerf}, GScream~\cite{wang2024learning}, GS-RoadPatching~\cite{chen2025gs}, and SplatFill~\cite{dahaghin2025splatfill}. Further, the recent 3DGIC~\cite{huang20253d} outperforms the state-of-the-art by using depth-guided cross-view consistency among multi-view images. 
The most recent concurrent work has pushed the quality even further.
GOR-IS~\cite{zhao2026gor} focuses on lighting consistency by decomposing the scene into intrinsic components and explicitly modeling light transport.
GP-GS~\cite{lee2026gpgs} uses a point cloud completion model, a coarse-to-fine inference strategy, image refinement, and a fine-tuning phase.
GaussFiller~\cite{ping2026gaussfiller} chooses the best view for inpainting with global-local alignment and then selects from multiple attempts the completion result with the highest semantic coherence.

\paragraph*{Score Distillation Sampling (SDS).} Similar to our approach, these methods generate splats using a 2D diffusion prior.
However, these are not directly applicable to completion as there is no mechanism to preserve existing splats out of the box. Dreamfusion~\cite{poole2022dreamfusion}, DreamGaussian~\cite{tang2023dreamgaussian}, and  MVDream~\cite{shi2023mvdream}, which uses a multi-view diffusion model that lifts 2D diffusion priors for 3D generation and makes them consistent using inflated 3D self attention, are just a few examples of the many methods available.
Similar to our approach, ComPC~\cite{huang2024compc} and SDS-Complete~\cite{kasten2023sdscomplete} use SDS to complete a 3D object, but handle only uncolored point clouds, rather than Gaussian splats.

%% file: sec/3_method.tex
\begin{figure}[t]
    \centering
    \includegraphics[width=\linewidth]{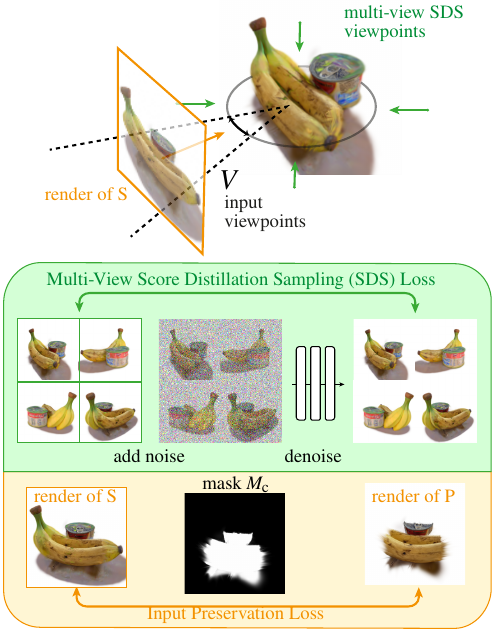}
    \caption{\textbf{Overview of \methodname}. The partial input $\PartialSplats$ is preserved while iterating with its masked ($M_c$) input preservation loss $\mathcal{L}_\text{p}$ and SDS loss $\mathcal{L}_\text{SDS}$ to complete the 3DGS object $\FullSplats$.}
    \label{fig:overview}
\end{figure}

\section{Method}

Given a partial 3D object represented by a set of Gaussian splats $\PartialSplats = \{\OneSplatPartial_1, \OneSplatPartial_2, \dots\}$, our goal is to generate a new set of Gaussian splats $\FullSplats = \{\OneSplatFull_1, \OneSplatFull_2, \dots\}$ that (i) represents a plausible completion of the 3D object, and (ii) exactly preserves the given part of the 3D object. An overview of our approach is shown in Figure~\ref{fig:overview}.

To generate a plausible object, we use a Score Distillation Sampling (SDS) approach based on MVDream~\cite{shi2023mvdream}. SDS allows us to generate plausible 3D objects using only a 2D diffusion prior, without having to rely on scarcely available 3D training data. We provide a short recap of this approach in Section~\ref{sec:object_generation}. As the SDS approach takes a text prompt $y$ of the desired completed object as additional input, we add $y$ to our inputs.

To preserve the given part of the 3D object, we keep the existing splats $\PartialSplats$ fixed during the SDS-based generation, and only optimize additional splats, effectively making sure that $\PartialSplats$ is a subset of $\FullSplats$. However, naively generating splats with this approach typically results in existing splats $\PartialSplats$ being progressively covered and hidden by the new splats in $\FullSplats$, as we show in our ablations. We believe this is due to a preference in the 2D prior for some types of objects that do not necessarily include completions of the given input. To address this problem, we propose a new \emph{input preservation loss} that makes sure that the partial input object is not hidden by new splats from a given distribution of viewpoints $V$. The viewpoints $V$ are given as input and could, for example, be based on a known capturing setup, or on user preferences. We describe this loss in Section~\ref{sec:input_preservation}.

Completing a 3D object with this approach requires suitable initialization and optimization strategies. In Section~\ref{sec:initialization_and_optimization}, we describe our approach that makes sure that new splats have sufficient coverage of regions that need to be completed without fully hiding existing splats.

In summary, given a partial 3D object as a set of splats, a text prompt, and a distribution of viewpoints $(\PartialSplats, y, V)$, respectively, we output the completed object as a set of splats $\FullSplats$ that are a superset of $\PartialSplats$ and preserve the appearance of $\PartialSplats$ from the viewpoints $V$.

\subsection{3D Object Generation with SDS}
\label{sec:object_generation}
Diffusion models~\cite{ho2020ddpm, song2021ddim, rombach2022latentdiffusion} generate images $x_0$ by iteratively denoising a full-noise image $x_T$ into increasingly less noisy versions $x_{T-1}, x_{T-2}, \dots, x_0$. In each step, a denoiser $g$ predicts a fully denoised version of the image $\hat{x}_0 = g(x_t, t, y)$, where $t$ is the time step and $y$ is a text prompt. The denoised image $\hat{x}_0$ is only a coarse approximation of $x_0$, but gives a good direction to follow towards the next less-noisy image $x_{t-1}$.

Score Distillation Sampling (SDS)~\cite{poole2022dreamfusion} uses the denoiser as a prior to generate 3D objects, by applying the denoiser to noised renders of the object:
\begin{equation}
    \mathcal{L}_{\text{SDS}} \coloneqq \mathbb{E}_{t,c}\ \| x_c - g\big(\epsilon(x_c, t), t, y\big) \|^2_2,
\end{equation}
where $x_c \coloneqq r(\FullSplats,c)$ is a differentiable render of the Gaussian splats $\FullSplats$ from viewpoint $c$ and $\epsilon(x,t)$ noises the image $x$ by an amount corresponding to timestep $t$.
This loss is used to iteratively optimize the splats $\FullSplats$.

We base our approach on MVDream~\cite{shi2023mvdream}, a variant of SDS that uses a denoiser fine-tuned to generate multi-view images of a 3D object:
\begin{equation}
\label{eq:mvsds}
\mathcal{L}_{\text{MVSDS}} \coloneqq \mathbb{E}_{t,c}\ \| x_c - g_\text{mv}\big(\epsilon(x_c, t), t, y\big) \|^2_2.
\end{equation}
Here $x_c \coloneqq r_\text{mv}(\FullSplats,c)$ is a multi-view image rendered from viewpoint $c$ and three additional viewpoints at the same elevation and equally spaced at 90 degree offsets along the azimuth. This multi-view formulation of SDS improves the consistency of the generated object across different views.


\subsection{Input Preservation Loss}
\label{sec:input_preservation}
To preserve the input splats $\PartialSplats$, we can include them in $\FullSplats$ and freeze them during the SDS optimization. This successfully preserves the splats, but we show in Section~\ref{sec:results} that the optimization tends to hide $\PartialSplats$ by covering it with other splats in $\FullSplats$. Note that a correct completion does need to cover the input splats from some viewpoints, for example, viewpoints showing the incomplete back side of the partial 3D object, but it should still be visible from other viewpoints $V$. The choice of viewpoints in $V$ -- independent of the above-mentioned SDS multi-views -- then defines, via their appearance, which splats will be preserved as visible. This may, for example, depend on the capturing setup used to obtain the input splats, or on an artistic choice, e.g., when using completion to ideate alternatives for 3D object parts, analogous to existing workflows in image editing~\cite{InvokeAI}.

We take this distribution of viewpoints $V$ as input and define a new loss that encourages the input splats $\PartialSplats$ to be visible from these viewpoints:
\begin{equation}
    \mathcal{L}_\text{p} \coloneqq \mathbb{E}_{c \sim V} \| M_c \cdot \big(r(\FullSplats,c) - r(\PartialSplats,c)\big) \|_2^2,
\end{equation}
where $\cdot$ denotes the element-wise product and $M_c \coloneqq r_a(\PartialSplats, c)$ is a 2D mask of the splats $\PartialSplats$ from viewpoint $c$, obtained as the alpha channel $r_a$ of the render $r(\PartialSplats, C)$. 

\subsection{Initialization and Optimization}
\label{sec:initialization_and_optimization}
Object completion is a task different than unconstrained generation, resulting in other dynamics of the SDS optimization. To adapt the SDS optimization to object completion, we carefully initialize the added Gaussian splats $\FullSplats \setminus \PartialSplats$ and define a schedule for the input preservation loss and for the amount of noise added to the renders in the SDS optimization.


\paragraph*{Initialization.}
Given the existing splats $\PartialSplats$ (20k-100k in our experiments), we initialize $|\FullSplats \setminus \PartialSplats| = 5000$ new Gaussian splats by uniformly distributing them in a sphere with diameter equal to the largest side of $\PartialSplats$'s bounding box and centered at the bounding box center.
We found that
in the first optimization steps, we can achieve faster and more stable convergence by encouraging splats to gather in regions that the missing parts of the object are likely to occupy. Intuitively, these missing parts are more likely to be visible from views other than those in $V$. Thus, we encourage new splats to initially gather in regions that are hidden from viewpoints in $V$. We remove the mask $M_c$ of the preservation loss in the first $n_\text{init} = 50$ iterations (setting $M_c$ to all-ones), effectively promoting renders from views in $V$ to match renders of the partial input object, thereby encouraging new splats to move behind the existing splats as seen from these viewpoints, or to reduce their opacity.

\paragraph*{Optimization.} 
In consecutive steps of the SDS optimization, we alternate between using the SDS loss $\mathcal{L}_\text{MVSDS}$ and a weighted sum of the SDS loss and the input preservation loss $\mathcal{L}_\text{MVSDS} + \lambda \mathcal{L}_\text{p}$, which reduces runtime without impacting quality. We use $\lambda = 2$ in our experiments and optimize for a total of $6000$ steps. Following the original Gaussian Splat implementation~\cite{kerbl20233d}, in the first $1500$ steps we subdivide the splats based on the gradient magnitude, giving us a total of $100$k - $200$k splats for $\FullSplats \setminus \PartialSplats$.
The diffusion time $t$ used in each optimization step (Eq.~\ref{eq:mvsds}) defines the amount of noise added to each image.
%
The amount of noise determines how much of the original rendering is preserved in the denoised image and how much is generated by the denoiser. Large amounts of noise only preserve coarse structures and encourage making larger changes to the Gaussian splat object, while small amounts of noise preserve everything but fine details and encourage making changes only to smaller details. In our experiments, we choose the following schedule for $t$ that reduces the noise level over time to focus only on refining details in later steps of the optimization:
\begin{itemize}
    \item Steps 1 to 3000:    $t \sim \mathcal{U}(0.4 T, 0.6 T)$ 
    \item Steps 3001 to 4500: $t \sim \mathcal{U}(20, 0.35 T)$ 
    \item Steps 4501 to 6000: $t \sim \mathcal{U}(20, 0.25 T)$ 
\end{itemize}
$\mathcal{U}$ denotes a uniform distribution and $T$ is set to 980. As discussed in Section 5 and shown in Table 2, our method is not sensitive to the choice of $t$ schedule and simpler schedules usually work as well.

%% file: sec/4_results.tex
\section{Results}
\label{sec:results}

We evaluate \methodname by comparing it to the most relevant prior work on our dataset of partial objects, showing that our method produces completions of similar or better quality, while preserving the existing partial objects much more accurately, followed by an ablation of our core technical contributions and a discussion of the main limitations.

\paragraph*{Metrics.}
Our goal is to produce completions of high plausibility while exactly preserving the partial input object. We measure \emph{plausibility} of the completion using the CLIP similarity $S_\text{CLIP}$ between a render of the partial input splats $\PartialSplats$ from one of the input viewpoints $V$ and a render of the completed splats $\FullSplats$ from the four viewpoints at azimuth offsets $0, 90, 180, 270$ from the input viewpoint. We measure \emph{input preservation} by comparing both color and depth renders of the partial splats $\PartialSplats$ from one of the input viewpoints $V$ to corresponding renders of the completed splats $\FullSplats$.
%
%
We compare with three metrics: the mean-squared error $E^\text{Color}_\text{MSE}$ and $E^\text{Depth}_\text{MSE}$ of color and depth, respectively, and the LPIPS~\cite{zhang2018perceptual} error $E_\text{LPIPS}$ of the color renders. Each of these is weighted by the mask $M_c$ to only focus on existing splats $\PartialSplats$.
%
%
We average all metrics over all objects of our dataset.


\paragraph*{Dataset.}
We evaluate on our new test set of partial Gaussian splat objects that we call {\scshape SplatComplete}. It consists of $39$ objects from different sources:

\noindent \emph{Multiview}: $24$ real-world objects were captured using using between $25$ and $50$ images taken from viewpoints distributed around the object. As a reconstruction method, we use the original Gaussian Splatting approach~\cite{kerbl20233d}. To obtain partial GS splat objects, we manually define a region to be preserved with a bounding box and discard all remaining splats. The remaining partial objects each comprise between $20$k and $100$k splats.

\noindent \emph{Singleview}: $5$ real-world objects were captured using a single image of the object. We use Depth Anything 3~\cite{depthanything3} to obtain a colored point cloud for the visible parts of the object and convert to Gaussian splats using a fixed splat size that is chosen manually for each object based on point cloud density.

\noindent \emph{LiDAR}: $5$ real-world objects were extracted from the Redwood Indoor Dataset \cite{Park2017}. We manually selected and cut the objects from room scans, removing clutter and any remaining background. Splat size was determined as in the Singleview subset.

\noindent \emph{Mesh}: $5$ synthetic objects were constructed from partially completed meshes obtained from Objaverse++ \cite{objaverse++}. We converted the mesh files to Gaussian splats with \emph{mesh2splat} \cite{scolari2025mesh2splat}.


\begin{table}[t]
\caption{\textbf{Comparison to Baselines.} We evaluate preservation of the input splats by comparing depth and color renders of the input splats from the input viewpoint to the corresponding renders of the completions. Plausibility of the completion is evaluated by measuring how similar the semantics of the input splats are to the semantics of the completion from multiple different viewpoints, as measured by the CLIP similarity.}
\label{tab:comparison}
\footnotesize 
\renewcommand{\arraystretch}{1.1}
\setlength{\tabcolsep}{5pt}
\begin{tabularx}{\linewidth}{r >{\centering\arraybackslash}X >{\centering\arraybackslash}X >{\centering\arraybackslash}X >{\centering\arraybackslash}X} 
     \toprule
     & \multicolumn{3}{c}{Input Preservation} & Plausibility  \\
     \cmidrule(l    r){2-4} \cmidrule(l){5-5}
     & $E^\text{Depth}_\text{MSE}\downarrow$ & $E^\text{Color}_\text{MSE}\downarrow$ & $E_\text{LPIPS}\downarrow$ & $S_\text{CLIP}\uparrow$ \\
     \midrule
     MVDream & 0.0167 & 0.2327 & 0.1561 & 66.985 \\
     Trellis MV & 0.0125 & 0.1548 & 0.1026 & 75.485 \\
     Trellis SV & 0.0116 & 0.1737 & 0.1050 & 75.478 \\
     InstantMesh & 0.0115 & 0.1205 & 0.0859 & 75.918 \\
     TripoSG & 0.0121 & 0.1641 & 0.0898 & 75.370\\
     \midrule
     \textbf{GSComplete (ours)} & \textbf{0.0084} & \textbf{0.0248} & \textbf{0.0370} & \textbf{77.944} \\
     \bottomrule
\end{tabularx}
\end{table}

%

\begin{figure*}[t!]
    \centering
    \includegraphics[width=\linewidth]{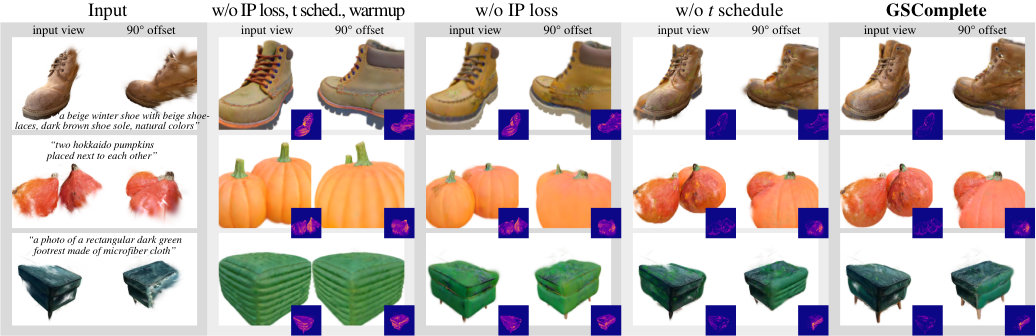}
    \caption{\textbf{Ablation.} We ablate several technical contributions, including the input preservation (IP) loss, and our optimization strategy, including warmup of the Gaussian splats in the first few iterations. For each object, we show two views of the partial input splats and the same two views for each ablation. Insets show errors in the preservation of input splats for each view as the difference between renders of the completed 3D objects and non-empty regions of the corresponding input view. We can see that our method preserves the input splats significantly more accurately than the baselines, with high plausibility of the completed shapes.} 
    \label{fig:ablation}
\end{figure*}

\paragraph*{Comparison.}
For \methodname, the user selects an input view $c$ and the distribution $V$ is then defined uniformly in range around $c$ with +/-50$^{\circ}$ azimuth. The 6000 SDS iterations take roughly $35$ minutes on a single RTX 3090 GPU.


A first baseline approach to GS completion is single-view reconstruction, by rendering the object from a viewpoint that is as complete as possible (the same viewpoint $c$ that we use as input for our method) and reconstruct the 3D object from the rendered image. \emph{Trellis}~\cite{xiang2025trellis}, \emph{InstantMesh}~\cite{xu2024instantmesh}, and \emph{TripoSG}~\cite{li2025triposg} are recent methods that cover this approach. We call Trellis used in this approach \emph{Trellis SV} to distinguish it from a multi-view version we describe next.

Single-view reconstruction misses any information about the partial input object that can't be captured by a single view. A different baseline approach thus renders multiple views of the partial object, inpaints any missing parts, and then reconstruct the 3D objects from the inpainted renders. We are not aware of any prior work that attempts this approach for 3D object completion, so we use our best attempt as a baseline: in addition to the viewpoint $V$, we render from two additional viewpoints with an azimuth offset of $90$ and $270$ degrees (we avoid $180$ degrees since from that viewpoint, the existing part of the object is typically fully hidden and thus the 2D inpainting method has no reliable reference to complete from that viewpoint), inpaint each view using SDXL~\cite{podell2023sdxl} with a mask based on the alpha channel of the rendered splats $\PartialSplats$, and finally use Trellis to reconstruct a 3D object from the inpainted multi-views. Since Trellis outputs 3D objects in a coordinate frame that is not aligned to the input image(s), we manually translate, rotate, and uniformly scale the models to best match the input splats. We call this multi-view baseline \emph{Trellis MV}.

A third baseline apporach is to use our SDS but without the input preservation loss or our changes to initialization and optimization. As we base our method on \emph{MVDream}~\cite{shi2023mvdream}, we run MVDream to only optimize the new splats $\FullSplats \setminus \PartialSplats$ while keeping existing splats $\PartialSplats$ fixed.



\begin{figure}[t]
    \centering
    \includegraphics[width=\linewidth]{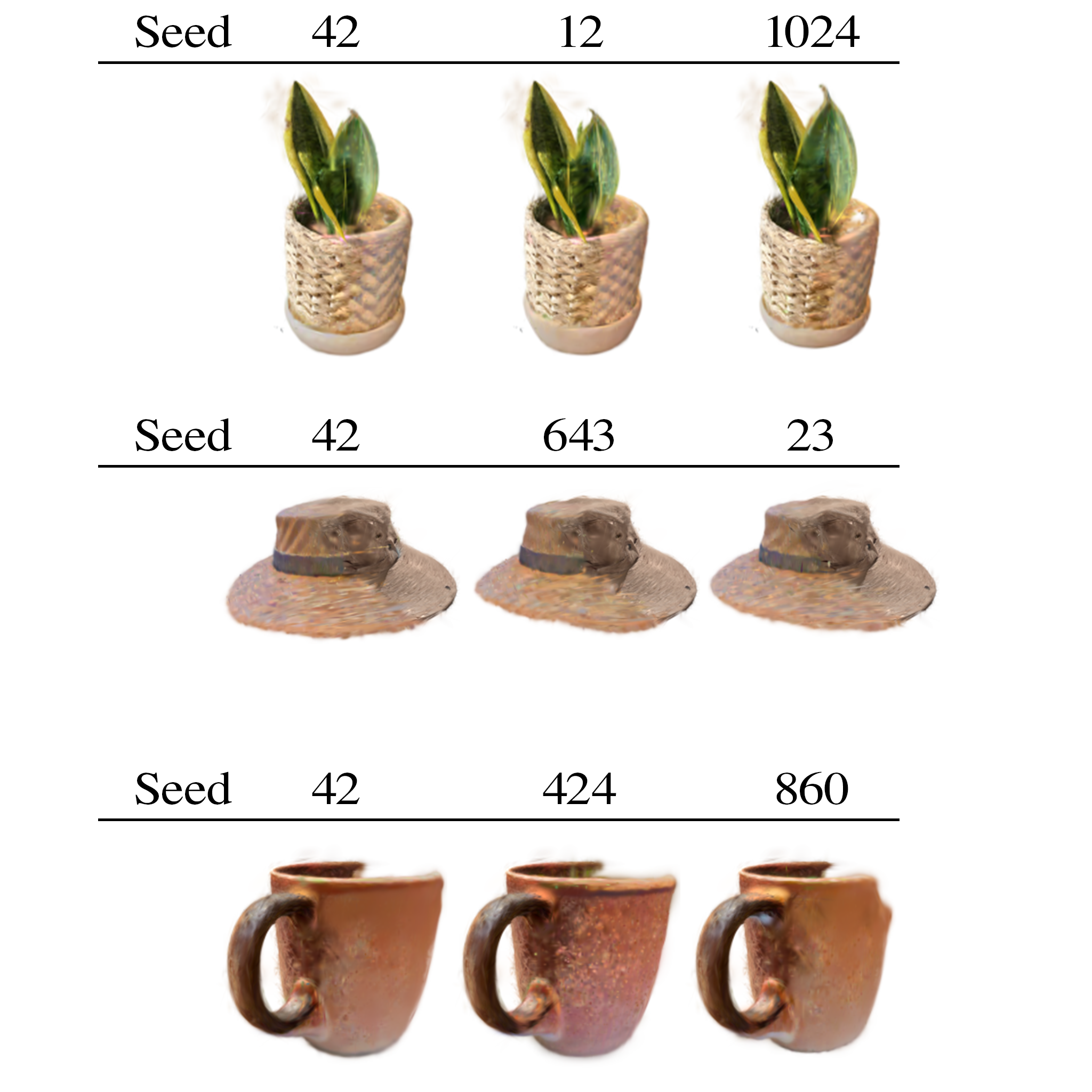}
    \caption{\textbf{Variance of Completions.} \methodname produces stable completions with different RNG seeds.} 
    \label{fig:variance_fig}
\end{figure}

\begin{figure*}[p]
    \centering
    \includegraphics[width=\linewidth]{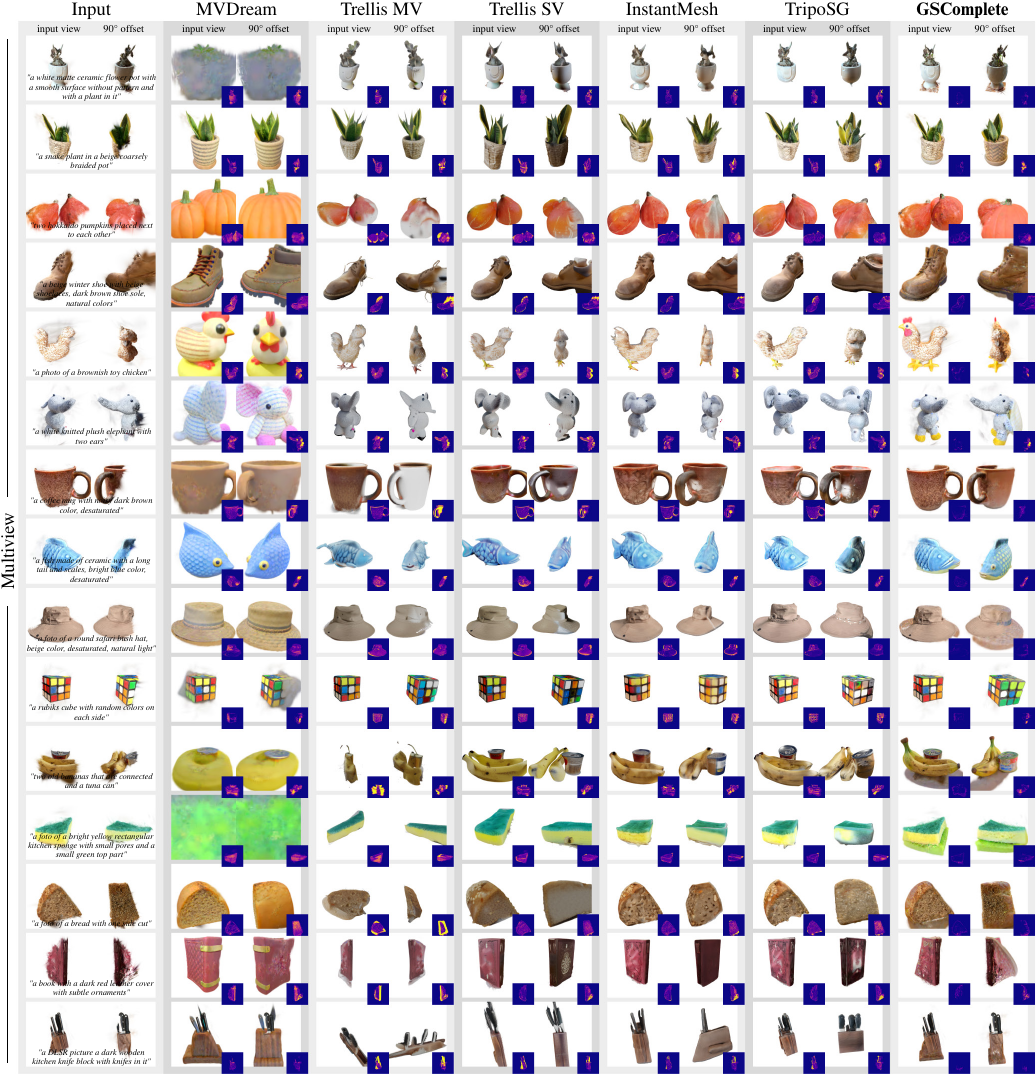}
    \caption{Qualitative results on partial GS objects obtained from multi-view images. We compare completions of \methodname to several baselines. For each object, we show two views of the partial input splats and the same two views for each completion. Insets show errors in the preservation of input splats for each view as the difference between renders of the completed 3D objects and non-empty regions of the corresponding input view. We can see that our method preserves the input splats significantly more accurately than the baselines, with high plausibility of the completed shapes.}
    \label{fig:comparison1}
\end{figure*}

\begin{figure*}[p]
    \centering
    \includegraphics[width=\linewidth]{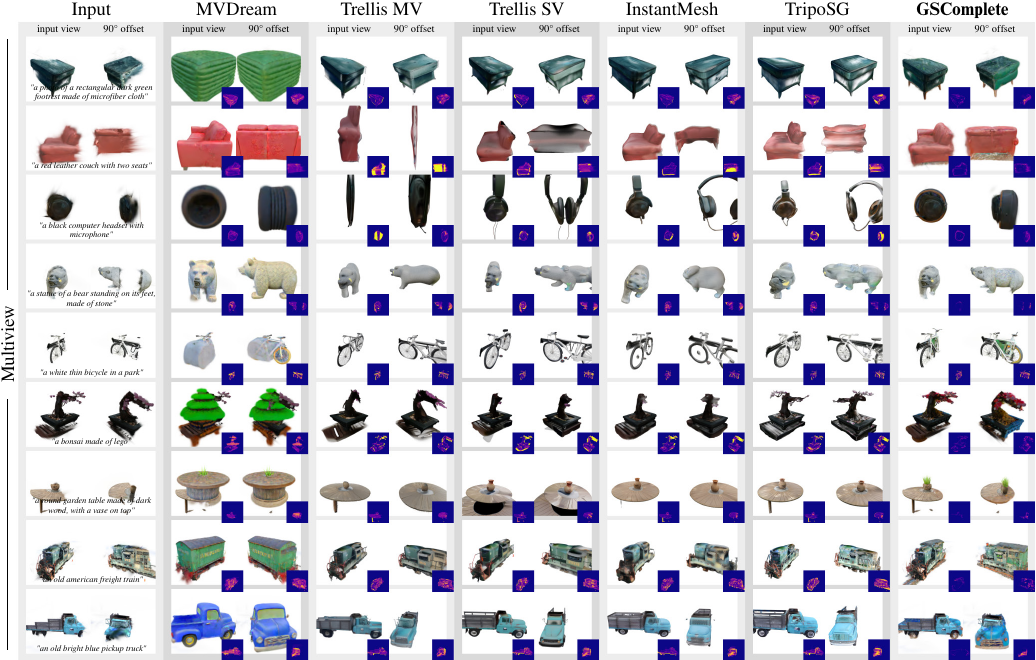}
    \caption{Additional qualitative results on partial GS objects obtained from multi-view images. We compare completions of \methodname to several baselines. For each object, we show two views of the partial input splats and the same two views for each completion. Insets show errors in the preservation of input splats for each view as the difference between renders of the completed 3D objects and non-empty regions of the corresponding input view.} 
    \label{fig:comparison2}
\end{figure*}

\begin{figure*}[p]
    \centering
    \includegraphics[width=\linewidth]{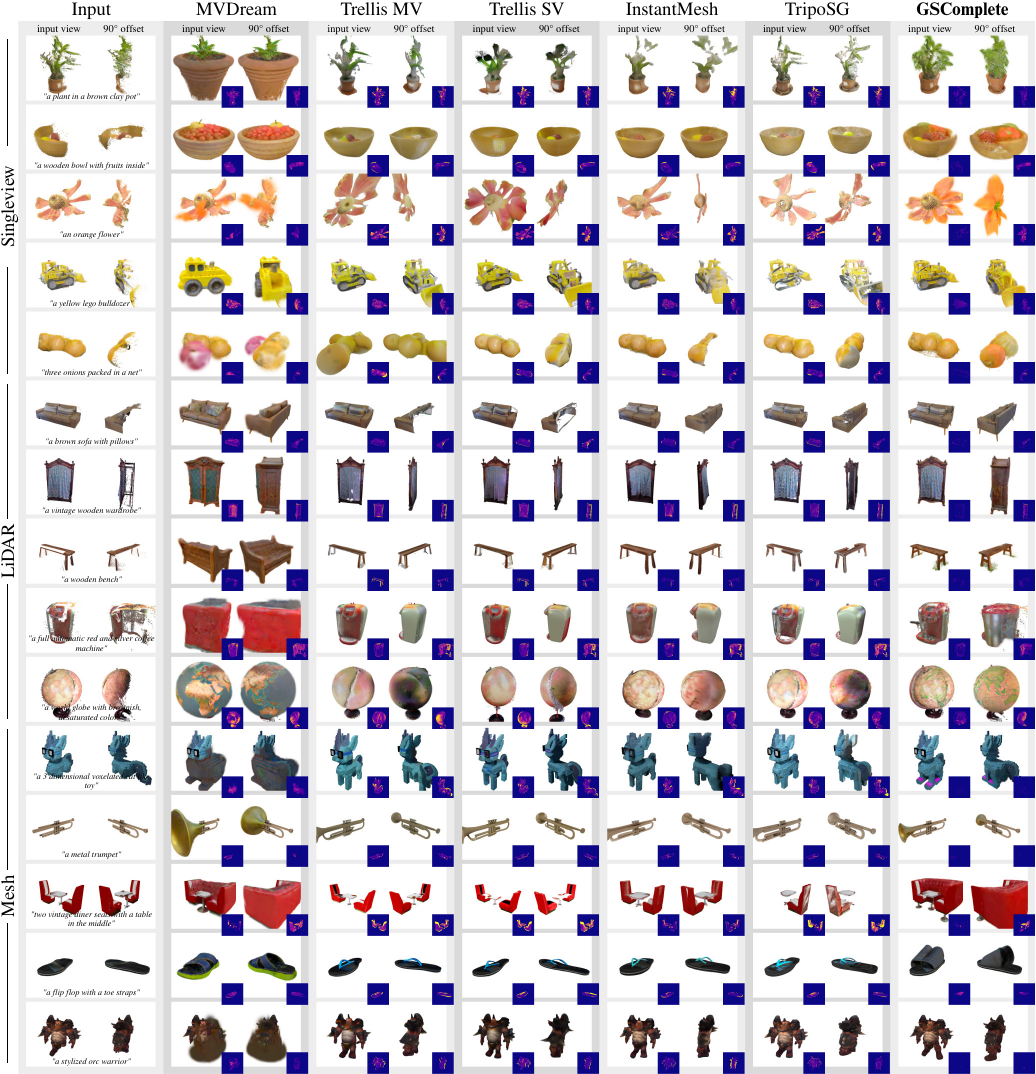}
    \caption{Qualitative results on partial GS objects obtained from single views, LiDAR scans, or partial meshes. We compare completions of \methodname to several baselines. For each object, we show two views of the partial input splats and the same two views for each completion. Insets show errors in the preservation of input splats for each view as the difference between renders of the completed 3D objects and non-empty regions of the corresponding input view.}
    \label{fig:comparison3}
\end{figure*}

A quantitative comparison is shown in Table~\ref{tab:comparison} and qualitative comparisons are given in Figures~\ref{fig:comparison1}, \ref{fig:comparison2}, and~\ref{fig:comparison3}. MVDream, lacking our input preservation loss, tends to cover existing splats, resulting in significantly different depths and colors when viewed from $V$, as reflected in the higher input preservation errors and lower clip similarity. Both Trellis SV and MV perform better, however, since they need to represent the partial input splats with image(s), the input's depth and color are preserved much less accurately, which also results in a slight deviation from the original semantics, as the lower CLIP similarity indicates. The 2D inpainting performed for multiple views in Trellis MV is not guaranteed to be view-consistent, introducing errors in the reconstruction. Completions from Trellis SV, InstantMesh, and TripoSG look reasonably accurate from the input views for several objects, as shown in Figures~\ref{fig:comparison1}, \ref{fig:comparison2}, and~\ref{fig:comparison3} (with InstantMesh having slightly lower quality than the other two), but the second view tends to show larger errors in input preservation. \methodname preserves the input splats accurately as seen from the input view. Differences to the input splats in other views mainly come from valid completions covering the input splats. We show in Figure~\ref{fig:variance_fig} that \methodname is not sensitive to different seeds for the random number generator.








\begin{table}[t]
\caption{\textbf{Ablation of core contributions.} We ablate several technical contributions, including the input preservation (IP) loss, and our optimization strategy, including warmup of the Gaussian splats in the first few iterations.}
\label{tab:ablation}
\footnotesize 
\renewcommand{\arraystretch}{1.1}
\setlength{\tabcolsep}{5pt}
\begin{tabularx}{\linewidth}{r >{\centering\arraybackslash}X >{\centering\arraybackslash}X >{\centering\arraybackslash}X >{\centering\arraybackslash}X} 
     \toprule
     & \multicolumn{3}{c}{Input Preservation} & Plausibility  \\
     \cmidrule(l    r){2-4} \cmidrule(l){5-5}
     & $E^\text{Depth}_\text{MSE}\downarrow$ & $E^\text{Color}_\text{MSE}\downarrow$ & $E_\text{LPIPS}\downarrow$ & $S_\text{CLIP}\uparrow$ \\
     \midrule
     w/o IP loss, $t$ schedule, warmup & 0.0167 & 0.2327 & 0.1561 & 66.985 \\
     w/o IP loss & 0.0124 & 0.1711 & 0.1272  & 70.489 \\
     w/o $t$ schedule & \textbf{0.0074} & \underline{0.0295} & \underline{0.0421}  & \textbf{78.233} \\
     \textbf{\methodname full} & \underline{0.0084} & \textbf{0.0248} & \textbf{0.0370}  & \underline{77.944} \\
     \bottomrule
\end{tabularx}
\end{table}

\paragraph*{Ablation Studies}


We ablate the core technical contributions of our approach on our dataset in Table~\ref{tab:ablation} and Figure~\ref{fig:ablation}. First, when removing the input preservation loss (w/o IP loss), the SDS optimization is free to fully cover the partial object $\PartialSplats$, effectively generating an object that is based on the given text prompt and ignoring the partial shape to a large extent. Examples are shown in Figure~\ref{fig:ablation}, columns 2 and 3. This is reflected in a higher input preservation error, as $\PartialSplats$ is no longer visible, and in lower plausibility of the completion, as the semantics of the completion are less similar to the partial input. Our approach is robust to the choice of the $t$ schedule, so replacing it with a simpler schedule where each optimization chooses $t$ uniformly in $[20,980]$ (w/o t schedule, adapted from DreamGaussian~\cite{tang2023dreamgaussian}) does not significantly impact the performance. Removing the initialization that warms up added splats with the unmasked input preservation loss (in addition to removing the t schedule and the IP loss) shows mainly a strong increase in the depth error $E^\text{Depth}_\text{MSE}$, as unpruned splats clutter the depth map.

\begin{figure}[t]
    \centering
    \includegraphics[width=\linewidth]{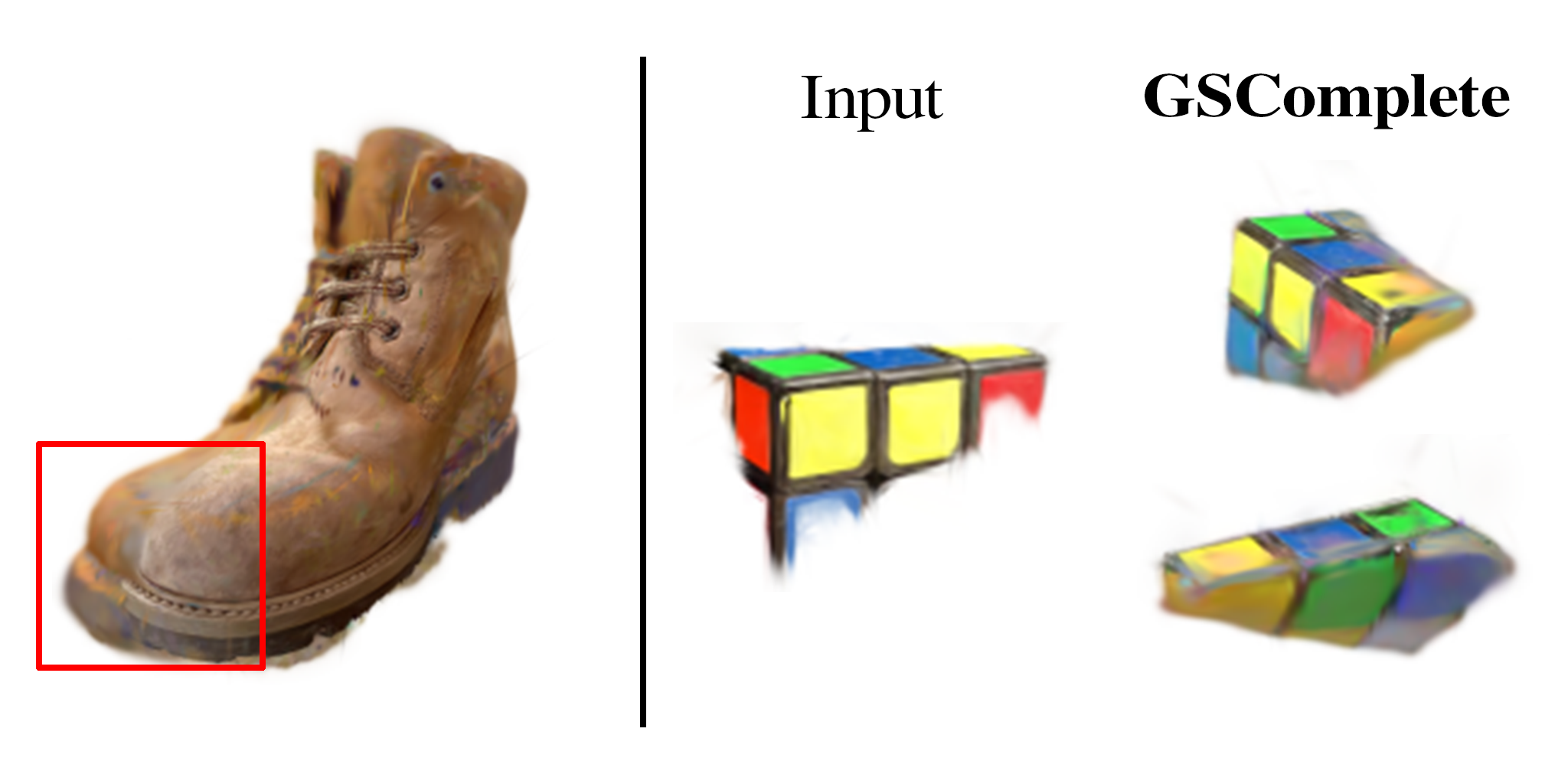}
    \caption{\textbf{Limitations of \methodname.} Left: texture and color seams. Right: Input with too little information.} 
    \label{fig:limitations_fig}
\end{figure}

\paragraph*{Limitations}

The quality of our result naturally depends on the richness of information from the input, even if our method manages to get good results from sparse input. If the input is too sparse, \methodname can have difficulties shaping meaningful information (see Figure \ref{fig:limitations_fig}).
The manually defined viewpoints/center-of-mass translations can easily be automated with heuristics, and the resolution of the diffusion output is currently limited to $256^2$ by MVDream's model architecture. Similar to other SDS methods, expansive scenes or scenes with complex backgrounds are difficult to handle, as they would require longer optimization and more careful camera placement. Like other SDS methods, we observe a slight amount of over-saturation due to the use of classifier-free guidance in the prior, sometimes creating visible color or texture seams, as shown in Figure \ref{fig:limitations_fig}.
Additionally, our implementation is not optimized, but its runtime is competitive with comparable methods such as SDS, MVDream, and 3DGIC.

%% file: sec/5_conclusion.tex
\section{Conclusion}

We revisited 3D object completion in the setting of Gaussian splats. We have shown that by introducing an input preservation loss to Score Distillation Sampling, we can complete partial objects represented as Gaussian splats in high quality while preserving the partial input more accurately than existing methods. Additionally, we introduced a test set of $18$ partial Gaussian splat objects captured from real-world objects that we use to evaluate our approach.

In future work, we plan to optimize the runtime of our method by using a faster diffusion algorithm and lowering resolution in the early stages, as well as using higher resolutions to further improve the output. Recomputing our preservation loss less frequently, but with a higher weight, can accelerate our method without losing quality. Iterating longer and densifying again later can address the color/texture seams, and adjusting hue and saturation values of the Gaussian splats could directly improve the texture quality even further. Finally, detecting multiple separate (partial) objects in a scene using heuristics and auto-generating prompts for these will allow feeding the objects individually to the diffusion module and thus effectively permitting completion of complex scenes.

%% file: sec/6_acknowledgements.tex
\section{Acknowledgements}
Special thanks to Pedro Hermosilla Casajus for important discussions on the project.

"This work has been partially funded by Wiener Wissenschafts-, Forschungs- und Technologiefonds (WWTF) projects ICT19-009 and ICT22-055, and FWF project PAT9402024."